\documentclass[11pt]{article}

\usepackage[final]{acl}

\usepackage{times}
\usepackage{latexsym}
\usepackage{soul,color}
\usepackage[T1]{fontenc}

\usepackage[utf8]{inputenc}

\usepackage{microtype}

\usepackage{inconsolata}

\usepackage{graphicx}

\usepackage{listings}
\usepackage{enumitem}
\usepackage{microtype}

\setlist[description]{style=multiline, leftmargin=0pt, labelsep=0.5em, itemsep=0.6\baselineskip}

\usepackage[most]{tcolorbox}

\newtcolorbox{promptbox}[1][]{%
  enhanced,
  breakable, 
  colback=white,    
  colframe=orange!80!white,   
  boxrule=0.8pt,    
  arc=2pt,          
  fonttitle=\bfseries, 
  title=#1          
}

\newtcolorbox{promptboxnb}[1][]{%
  colback=white,    
  colframe=orange!80!white,   
  boxrule=0.8pt,    
  arc=2pt,          
  fonttitle=\bfseries, 
  title=#1          
}

\usepackage{minted}

\usepackage{quoting}
\quotingsetup{vskip=0pt,leftmargin=0pt,rightmargin=0pt}

\usepackage{dirtree}

\usepackage{booktabs}
\usepackage{tabularx}
\usepackage{pdflscape}
\usepackage{array}

\title{Stakeholder Suite: A Unified AI Framework for Mapping Actors, Topics and Arguments in Public Debates}

\author{
Mohamed Chenene$^{1}$ \quad Jeanne Rouhier$^{1}$  \quad Jean Daniélou$^{2}$ \quad
Mihir Sarkar$^{3}$  \quad Elena Cabrio$^{4}$\\
$^{1}$ENGIE Lab CRIGEN, France \\
$^{2}$Centre de Sociologie de L’Innovation, CNRS, UMR 9217, Mines ParisTech, PSL University \\
$^{3}$ENGIE Research \& Innovation, France \\
$^{4}$Université Côte d’Azur, CNRS, INRIA, I3S, France \\
\texttt{mohamed.chenene1@gmail.com}, \ \texttt{jeanne.rouhier@external.engie.com}, \\ \texttt{jeandanielou@live.fr}, \ \texttt{mihir.sarkar@engie.com}, \ \texttt{elena.cabrio@univ-cotedazur.fr}
}

\begin{document}
\maketitle

\begin{abstract}
Public debates surrounding infrastructure and energy projects 
involve complex networks of stakeholders, arguments, and evolving narratives. Understanding these dynamics is crucial for anticipating controversies and informing engagement strategies. 
This paper presents \textbf{Stakeholder Suite}, a framework deployed in operational contexts for mapping actors, topics, and arguments within public debates. The system combines actor detection, topic modeling, argument extraction and stance classification in a unified pipeline. Tested on multiple energy infrastructure projects as a case study, the approach delivers fine-grained, source-grounded insights while remaining adaptable to diverse domains. 
The framework achieves strong retrieval precision and stance accuracy, producing arguments judged relevant in 75\% of pilot use cases. Beyond quantitative metrics, the tool has proven effective for operational use: helping project teams visualize networks of influence, identify emerging controversies, and support evidence-based decision-making.
\end{abstract}

\section{Introduction}

In the era of social media, public controversies can emerge rapidly and significantly affect the reputation and operations of organizations. Controversy analysis aims to systematically track and interpret such public debates by identifying the stakeholders involved, their relationships, their stances and the evolution of discourse over time. A structured understanding of these dynamics enables organizations to anticipate conflicts, adapt communication strategies, and mitigate potential delays or financial losses.  
In the energy sector, large-scale renewable infrastructure projects (e.g., wind farms, solar plants, district heating, hydrogen, or ammonia transport) frequently trigger public consultations and debates \cite{LAPATIN2023113507}. Numerous stakeholders (citizens, associations, local officials and activists) express their views across diverse digital channels, making the monitoring and analysis of these discussions both complex and time-consuming for project developers.

This work addresses the question of how 
computational methods can support and enhance a transparent analysis of public debates. Our objective is to develop a \emph{generic and adaptable framework} capable of mapping stakeholders, identifying debated topics, and extracting arguments and stances to provide an exhaustive and objective view of controversial debates.
We introduce \textbf{Stakeholder Suite}, that combines Large Language Models and Retrieval-Augmented Generation (RAG) \cite{rag} to build a structured representation of public debates. LLMs \cite{flan, fewshots, llm-argument} now perform effectively 
for various NLP tasks, enabling fast deployment of complex analysis pipelines without large annotated datasets. RAG improves factual grounding and transparency by coupling generation with evidence selection, mitigating hallucinations, and improving traceability \cite{hallucination}.  

\noindent This paper makes the following contributions:
    1. We present a transparent and solid framework for the analysis of public debates, which jointly models stakeholders, topics, and arguments within a unified data structure;
    2. We propose a fine-grained approach to topic modeling and argument analysis that leverages LLM-based reasoning, enabling richer and contextual insights. 

To date, the system has been deployed across more than ten renewable energy projects, offering project teams actionable insights into local dynamics and stakeholder perceptions. To the best of our knowledge, this is among the first operational frameworks that combines stakeholder mapping, fine-grained argument extraction, and topic modeling within a single and updatable pipeline spanning heterogeneous texts. A demo video is available.\footnote{\url{https://youtu.be/c1b6QgyVkws}}



\section{Related Work}
\label{sec:related_work}

Mapping stakeholders and their stances across public debates intersects with several markets and research areas: media intelligence, social listening, public affairs, controversy analysis, policy analysis and argument mining (AM). 

\paragraph{Commercial solutions.} Social listening and sentiment analytics platforms as Talkwalker, Brandwatch or Meltwater, aggregate large-scale social and web data, offering dashboards for trends, topics, influencer identification, and sentiment. A second line of commercial tools centers on network views of entities, topics, and audiences. NetBase Quid and Pulsar Platform provide interactive graphs, topic clustering, community analysis, and audience segmentation across social and news sources. These capabilities help reveal ecosystems (e.g., co-mention networks and narrative clusters) and identify influential accounts. The Stakeholder Company (TSC.ai) maintains a large stakeholder repository with influence mapping and engagement workflows and comes closest to stakeholder-centric use cases in practice. Public affairs and government-relations platforms provide structured data and workflows for legislative and regulatory engagement. Quorum integrates bills, hearings, and institutional actors across multiple jurisdictions, supporting monitoring, alerting, and outreach. Some products also incorporate retrieval-augmented assistants for quick knowledge access. While effective within institutional policy contexts, these systems are not intended as general-purpose frameworks across heterogeneous public debates.\footnote{Commercial platforms discussed: Talkwalker (\url{https://www.talkwalker.com}), Brandwatch (\url{https://www.brandwatch.com}), Meltwater (\url{https://www.meltwater.com}), NetBase Quid (\url{https://www.quid.com}), Pulsar Platform (\url{https://www.pulsarplatform.com}), TSC.ai (\url{https://tsc.ai}), Quorum (\url{https://www.quorum.us}).}

\paragraph{Academic research.} Foundational work in controversy analysis \cite{controversymap} establishes sociological methods for systematically mapping controversies and constructing networks from debates. 
Automated policy analysis applies NLP pipelines to structure regulatory content, extracting measures, actors, and impacts from environmental and institutional documents \cite{forestpolicy, climatepolicy}. For instance, \cite{beyondnlp}'s framework is a knowledge graph-oriented approach that allows rapid review of policy documents through entity search, topic analysis, and policy search. While these approaches enable efficient document analysis, they typically operate on formal policy texts rather than public debate corpora.
AM offers complementary methods for analyzing contested discourse, including claim detection, stance classification, and argument clustering \cite{ArgumenText, ibmdebater} 
in different kinds of structured contexts: essays, online debate platforms, legal documents, and political debates. Political debate analysis, which examines clash points, strategies, and argumentative structures \cite{debatvis, conch, disputool}, shares similarities with energy infrastructure debates in involving political actors, citizen associations, and multi-channel information flows.

While prior work provides valuable components (e.g.,  large-scale data access, influencer and network analytics, institutional monitoring, and argument mining methods), an end-to-end production-oriented approach that jointly maps stakeholders, detects and organizes topics, and extracts source-grounded arguments with stance across heterogeneous debate corpora has not yet been proposed. This gap motivates the development of retrieval-augmented frameworks adaptable across domains where public debate and stakeholder engagement are central.

\section{Methodology}
\label{sec:methodology}

\begin{figure*}
    \centering
    \includegraphics[width=16cm]{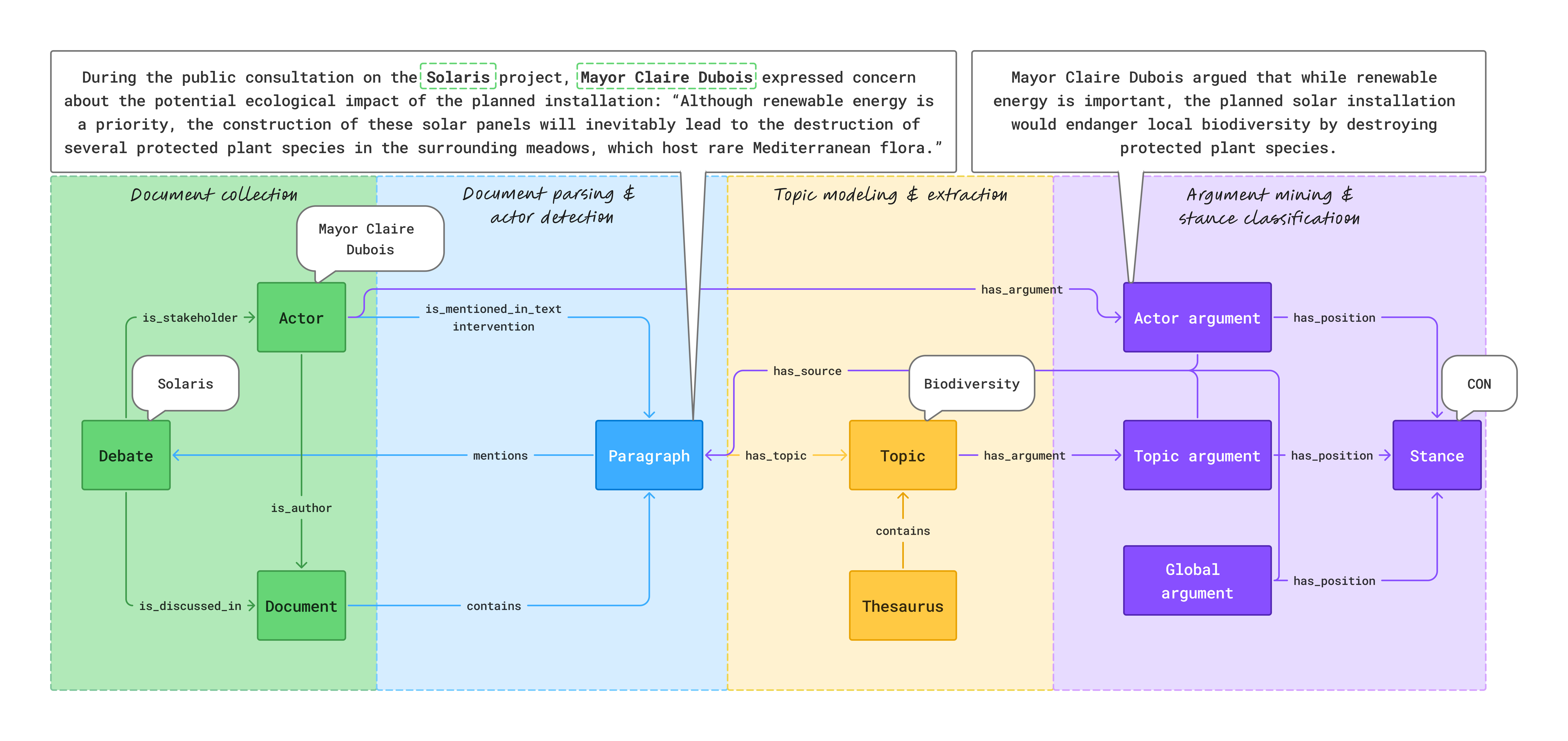}
    \caption{Blueprint of the \textbf{Stakeholder Suite} framework with an actor argument example: (1) Construction of a database containing stakeholders and documents related to the public debate; (2) Parsing of documents into paragraphs and detection of actor mentions and interventions within the text; (3) Linking of paragraphs to topics from the thesaurus using semantic similarity; (4) Generation of arguments through a RAG pipeline along three analytical axes (global, topic-specific, actor-specific) together with automatic stance classification for each argument.}
    \label{fig:placeholder}
\end{figure*}

The goal of \textbf{Stakeholder Suite} (Figure \ref{fig:placeholder}) is to provide a comprehensive snapshot of a public debate. Its components are designed to address three main questions:
(1)  \textit{Who is speaking?} \textrightarrow\ Actor detection;
(2)    \textit{What are they talking about?} \textrightarrow\ Topic modeling and extraction;
(3)   \textit{What is their position?} \textrightarrow\ Argument mining and stance classification.


\subsection{Data Collection}

To start, two main components are required:

\noindent    (1) \textbf{Stakeholder list:} manually or automatically identified actors participating in the debate;
    
\noindent    (2) \textbf{Document database:} a corpus integrating diverse textual sources on the debate, including:
    (a) \textbf{Debate-centric data:} press articles, policy papers and administrative documents, typically obtained through data brokers, web scraping or manually uploaded;
    (b) \textbf{Actor-centric data:} social media posts, websites and blogs associated with listed stakeholders.

\paragraph{Data quality.} Both data types are essential for ensuring that stakeholders are properly represented within the corpus. 
The most informative argumentative content often originates from stakeholder-generated material. Based on our experience in the energy sector, high-yield argumentative sources include public debate transcripts, stakeholder blog posts, and social media statements. Conversely, press and institutional documents tend to be less argumentative but provide valuable contextual information for topic modeling.

\subsection{Data parsing and actor detection}
\textbf{Data parsing.} After constructing the database, the next step is to structure it and establish semantic links between entities. Given that sentences are often too short to capture full meaning, while documents can vary greatly in length, we adopt the paragraph as the atomic analysis unit, as it provides a balance between contextual coherence and processing granularity. 
%
%
In our implementation, paragraph segmentation, coreference resolution, and actor speech extraction are handled jointly through a single LLM prompt (see Appendix~\ref{appendix:preprocessing}). All paragraphs are embedded and indexed in OpenSearch vector database for full-text and semantic search.

\textbf{Actor detection.} Once paragraphs are extracted, we link them to the corresponding actors from the stakeholder list. We define three relation types:

\noindent    (1) \textit{is\_author:} the actor of the document (e.g., blog post, tweet), known from the actor-centric corpus.
    
    \noindent (2) \textit{is\_mentioned\_in\_text:} explicit textual mentions of an actor in a paragraph. To improve the precision for common or ambiguous names, we search for direct matches using the known actor list and verify them with FLAIR NER~\cite{flair}.
    
\noindent     (3) \textit{intervention:} actor speech segments in debate transcripts. Interventions are more valuable content than mentions because they report the actor's speech. 
We use the LLM parser mentioned above.

Mentions of the debate are also detected at the paragraph level using a predefined list of aliases. This step is necessary because, even if a document refers to the debate as a whole, not all paragraphs within it necessarily discuss the debate directly.

\subsection{Topic modeling and extraction}

With a structured database of paragraphs linked to actors, the next step is to identify the topics discussed. Applying standard unsupervised clustering to highly domain-specific corpora often produces generic or irrelevant topics. 
Following \cite{clustering}, 
we leverage LLMs to construct a domain-specific thesaurus: a structured list of topics and their corresponding descriptions. The process involves the following steps:

 \noindent   (1) \textbf{Corpus chunking:} split the database into coherent subsets according to the debate characteristics. For long-running cases, use temporal windows (e.g., monthly or quarterly); for political debates, segment by speaker or party.
    
    \noindent (2) \textbf{Chunk description via RAG:} for each subset, run a RAG with a query such as: \textit{``What are the emerging topics related to \{debate\_name\}?''}, and extract the resulting topic candidates.
     
    \noindent (3) \textbf{Topic clustering:} cluster the generated topics using methods such as HDBSCAN or Spectral Clustering based on a semantic affinity matrix \cite{hdbscan, spectral-clustering}.
    
   \noindent (4) \textbf{Cluster summarization:} apply an LLM to name each cluster and provide representative examples, producing two descriptive levels (topic and subtopic) to preserve granularity.
    
   \noindent (5) \textbf{Thesaurus aggregation:} run a final LLM pass to consolidate results into a unified thesaurus containing topics, descriptions, and subtopics.

Although this pipeline heavily relies on LLMs, it has proved most effective for generating high-resolution topic descriptions. Moreover, analyzing topics across temporal slices reveals trends and weak signals of business relevance. Empirically, naive clustering methods consistently failed to capture debate-specific nuances, whereas our LLM-augmented approach produced coherent, non-generic topics. Appendix \ref{appendix:topic} provides an example of a thesaurus built with this approach and the prompts used for steps 4 and 5.
Once the thesaurus is defined, topic occurrences are computed across the corpus. We compute cosine similarity between topic descriptions and paragraph embeddings, retaining matches above a predefined threshold ($t=0.15$) to balance quality and quantity of paragraphs per topic. When multiple topics qualify, the top three by similarity are selected. 

\subsection{AM and stance classification}

We adopt a RAG architecture for argument detection, because (i)
  it is easily implemented in production environments through LLM API calls (in particular, we use \texttt{gpt-4o-mini} \cite{openai2024gpt4ocard} on the Azure platform, selected for compliance and cost-quality balance);  (ii) it enables the generation of concise, one-sentence arguments that are easily interpretable by end users; (iii) it preserves traceability by maintaining explicit links between generated outputs and their source documents.

A known limitation of RAG, however, is its non-exhaustive coverage of the corpus, as it retrieves only the top-$K$ relevant passages prior to generation. To mitigate this, we partition the database into coherent chunks and query each subset independently, allowing the extraction of more than 300 arguments per debate. This strategy achieves a practical balance between corpus coverage, computational cost, and analytical depth.

\subsubsection{Temporal and Dimensional Splits}

To maximize coverage of the public debate and reveal dynamics, we partition the corpus along two axes: (i) \textbf{time}, to trace how arguments evolve; and (ii) \textbf{dimension}, to focus on \textit{stance}, \textit{actors}, or \textit{topics} derived from previous steps. We operationalize this with three uniform query families.

\paragraph{(1) Global argument.}
\textbf{Goal:} enumerate debate arguments \emph{in favor of} or \emph{against} for each year.  

\textbf{Database filter (pseudo-Cypher):}
\begin{quoting}
\small{\texttt{MATCH (p:PARAGRAPH) WHERE p.date.year = \$year RETURN p}}
\end{quoting}

\textbf{Prompt to LLM:} \textit{``What are all the arguments \{in favor/against\} \{debate\_name\}?''}

\textbf{Rationale:} yearly stance snapshots surface trends and turning points.

\paragraph{(2) Actor argument.}
\textbf{Goal:} extract arguments attributable to a specific stakeholder, prioritizing paragraphs that are most on-topic for the target debate (no temporal split due to data sparsity).

\textbf{Database filter (pseudo-Cypher):}
\begin{quoting}
\small
\texttt{(a:ACTOR)-->(p:PARAGRAPH)-[:MENTIONS]->(d:DEBATE)        RETURN p AS p\_debate}\\
\texttt{(a:ACTOR)-->(p:PARAGRAPH)-[:HAS\_TOPIC]->(:TOPIC)        RETURN p AS p\_topic}\\
\texttt{(a:ACTOR)-->(p:PARAGRAPH)                                RETURN p AS p\_all}\\[2pt]
\texttt{IF COUNT(p\_debate) > 25 THEN RETURN p\_debate}\\
\texttt{ELSE IF COUNT(p\_topic)  > 25 THEN RETURN p\_topic}\\
\texttt{ELSE RETURN p\_all}
\end{quoting}

\textbf{Prompt to LLM:} \textit{``What are the arguments of the stakeholder \{actor\_name\} about \{debate\_name\}?''}

\textbf{Rationale:} We prioritize \emph{debate-specific} paragraphs for precision; if these are insufficient (\(\leq 25\)), we fall back to \emph{topic-linked} paragraphs, and otherwise include \emph{all} paragraphs for coverage. The threshold was determined empirically from a study of more than 150 actors and was optimized to maximize the number of stances extracted.

\paragraph{(3) Topic argument.}
\textbf{Goal:} collect arguments tied to a specific topic, per year, decoupled from the debate stance if necessary.  

\textbf{Database filter (pseudo-Cypher):}
\begin{quoting}
\small{
\texttt{MATCH (p:PARAGRAPH)-[:HAS\_TOPIC]\textrightarrow (t:MACRO\_TOPIC)}\\
\texttt{WHERE p.date.year = \$year AND t.name = \$topic\_name}\\
\texttt{RETURN p}
}
\end{quoting}
\textbf{Prompt to LLM:} \textit{``What are the arguments related to the topic \{topic\_name\}?''}

\textbf{Rationale:} topics may drift from debate framing; isolating them improves thematic resolution. It can also give a broader vision of the subject.

For a setup with $\sim$50 actors, 8 topics, and a 5-year span, and assuming $\sim$3 arguments are generated per query we get approximately 300 arguments per debate.
Combining temporal and dimensional partitions with standardized prompts yields a scalable, interpretable, and trend-aware argument set while keeping retrieval focused and noise low.

\subsubsection{RAG Pipeline}

The RAG pipeline structures the final argument extraction process. For each query identifier, it operates in four sequential stages:

\noindent    (1) \textbf{Retrieval:} the database is pre-filtered according to the previously mentioned split. We retrieve the top-$K$ paragraphs ($K=25$) using Maximum Marginal Relevance \cite{mmr} with $\lambda = 0.8$ to balance relevance and diversity.
    
    \noindent    (2) \textbf{Argument generation:} the query and retrieved paragraphs are passed to the LLM, which returns a structured list of arguments, each explicitly linked to one or more source paragraphs. The output is then parsed using regex to extract the argument text and the corresponding source identifiers.
    
    \noindent    (3) 
    \textbf{LLM-as-a-Judge} \cite{llmasjudge}:  a second LLM performs an automatic quality check, verifying that each generated argument is supported by its sources and flagging potential hallucinations.
    
   \noindent    (4)
    \textbf{Stance classification:} each validated argument is assigned a stance label (\texttt{PRO}, \texttt{CON}, or \texttt{NEUTRAL}).

\noindent Appendix \ref{appendix:argument} shows the prompts for steps 2 to 4, together with examples of extracted arguments.

\subsection{Visualization: Stakeholder Mapping}

To visualize stakeholder relationships, we construct a connected network where each node represents an actor. Node color encodes stance (\texttt{green} = pro, \texttt{red} = con, \texttt{grey} = neutral), and node size reflects the frequency of debate mentions during actor interventions. Edges are drawn when one actor explicitly references another within an intervention, capturing direct interactions and influence links. An example can be found in Appendix \ref{appendix:app}.



\section{Evaluation}
\label{sec:evaluation}
As previously described, our framework addresses three main tasks, i.e., NER for actor detection, semantic text similarity for topic extraction and RAG for argument generation. Since for the first two we rely on standard, validated approaches \cite{flair-ner, mteb}, in the following we focus on the RAG pipeline, with emphasis on retrieval, generation, and stance classification. 

\subsection{Retrieval Evaluation}

Retrieval performance was assessed across eight energy projects for both \texttt{PRO} and \texttt{CON} stances (16 queries in total). Each evaluation mixed relevant and randomly sampled paragraphs to approximate realistic corpus conditions, producing datasets of 200 candidates per query, with an average of about 35 relevant documents. 
We report Precision@K as the primary metric, focusing on diversity–relevance trade-offs using Maximum Marginal Relevance (MMR). As shown in Table~\ref{tab:lambda_retrieval}, optimal performance occurs for $\lambda \in [0.7, 0.8]$, confirming that moderate diversity does not deteriorate retrieval quality. The mean Precision@20 reaches 0.59, with variability driven by project-specific content density, sufficient for reliable argument generation downstream.

\begin{table}[ht!]
    \centering \small
    \begin{tabular}{cccccc} 
        \hline
        \textbf{MMR $\lambda$} & \textbf{Precision@10} & \textbf{Precision@20} \\ \hline
        0.5    & 0.41$\pm$0.13  & 0.40$\pm$0.09   \\ 
        0.6    & 0.59$\pm$0.17  & 0.55$\pm$0.14   \\ 
        0.7    & \textbf{0.64$\pm$0.20}  & 0.57$\pm$0.18  \\ 
        0.8    & 0.64$\pm$0.25  & \textbf{0.59$\pm$0.17} \\ 
        0.9    & 0.63$\pm$0.26  & 0.58$\pm$0.19  \\ 
        1.0    & 0.63$\pm$0.26  & 0.58$\pm$0.19  \\ \hline
    \end{tabular}
    \caption{\label{tab:lambda_retrieval}
    Retrieval performance across different $\lambda$ values. 
    Metrics are reported as mean $\pm$ standard deviation. 
  }
    \label{tab:lambda_performance}
\end{table}

\begin{table}[ht!]
  \centering \small
  \begin{tabular}{lllll}
    \hline
    \textbf{Component} & \textbf{Class} & \textbf{Precis.} & \textbf{Recall} & \textbf{F1} \\
    \hline
    \textbf{LLM-as}
      & BAD & 0.42 & 0.75 & 0.54 \\
     \textbf{a-Judge} & GOOD & 0.77 & 0.45 & 0.57 \\
      & \textbf{Macro avg} & \textbf{0.59} & \textbf{0.60} & \textbf{0.55} \\
    \hline
    \textbf{Stance}
      & AGAINST & 0.96 & 0.93 & 0.94 \\
     \textbf{Classification} & NEUTRAL & 0.69 & 0.76 & 0.73 \\
      \textbf{(ACTOR)} & PRO & 0.66 & 0.60 & 0.63 \\
      & \textbf{Macro avg} & \textbf{0.77} & \textbf{0.76} & \textbf{0.77} \\
    \hline
    \textbf{Stance}
      & AGAINST & 1.00 & 0.61 & 0.76 \\
      \textbf{Classification} & NEUTRAL & 0.72 & 0.89 & 0.80 \\
     \textbf{(TOPIC)} & PRO & 0.78 & 0.70 & 0.74 \\
      & \textbf{Macro avg} & \textbf{0.83} & \textbf{0.73} & \textbf{0.76} \\
    \hline
  \end{tabular}
  \caption{\label{tab:llm_position}
    Performance of the LLM-as-a-Judge and Stance Classification modules. 
  }
\end{table}

\subsection{LLM-as-a-Judge: Generation Evaluation}

To assess generation quality, a secondary LLM evaluated each argument-source pair to identify unsupported or hallucinated outputs. Our primary objective was to minimize false positives visible to end users, prioritizing a low error rate over exhaustive recall. In practice, this meant maximizing \texttt{BAD} argument recall while maintaining acceptable coverage of \texttt{GOOD} ones, to preserve user trust and prevent misinterpretation of the system’s outputs. 
Several prompt configurations were tested to balance strictness and recall. On a validation set of 739 samples, the best-performing validator achieved high \texttt{BAD} recall to protect users from unsupported claims, accepting some loss of \texttt{GOOD} recall (Table~\ref{tab:llm_position}).

\subsection{Stance Classification}

Following the cross-topic argument classification approach proposed by \cite{cross-topic-argument}, 3 annotators with expertise in Computational Linguistics annotated 350 arguments with their true stance toward one energy project to evaluate LLM classification performance. As shown in Table~\ref{tab:llm_position}, we report the overall performance. For reference, \cite{stance-classification} estimates human performance on cross-topic stance classification at $0.81$ (F1-score). Our results fall close to this upper bound, indicating reliable performance for our application.



\section{A Case Study for Users Evaluation}
\label{sec:use_case}



User evaluations globally demonstrate that the system’s analytical outputs are both accurate and operationally valuable (Appendix \ref{appendix:app} shows screenshots of the application). The application was used by 10 end-users across 9 different energy pilots. During onboarding, roughly 600 documents were processed per project, followed by about 20 new documents per month for continuous monitoring. Users particularly valued the argument summaries and mapping visualizations for improving internal communication and evidence-based engagement planning (\textit{“Stakeholder Suite gives us a territorial radar. It saves us days of reading and helps us get straight to what really matters.”}, from a pilot user's notes). In parallel, we directly collected their feedback through the application. On a sample of approximately 200 arguments, 75\% were judged relevant or useful by project teams.  

In the following, we present some insights gathered with the Stakeholders Suite on a solar project.
Located in southern France, the \emph{Montagne de Lure} area has experienced several solar development projects, some of which have triggered tensions among local associations and residents. Using Stakeholder Suite, project teams aggregated data from past and ongoing debates to identify active opponents, supportive officials, and recurring public concerns such as deforestation, landscape alteration, and threats to protected species.


\noindent \textbf{Topic Extraction.}
Automatic clustering surfaced two dominant themes, i.e., land use conflicts and regulation and participation.  
The land-use cluster revealed strong attention to biodiversity and landscape protection, with repeated mentions of the ocellated lizard, a locally protected species. This insight led the project team to initiate additional environmental assessments before site validation.  
The regulation cluster showed that many objections targeted the lack of transparency and citizen participation rather than the solar technology itself, prompting stronger consultation efforts. A third cross-topic finding indicated a clear preference for solar installations on brownfields or rooftops instead of forests or farmland.

\noindent \textbf{Argument Analysis.}
AM and stance classification exposed a well-structured opposition led by environmental groups. Their discourse gained visibility through local media and a published book that became a symbolic reference against solar projects in forested areas.  
Conversely, arguments from supportive actors (including other project owners) highlighted the site’s strong solar potential and its strategic alignment with EU renewable goals, offering communication models for future initiatives.

\noindent \textbf{Network Insights.}
It revealed tightly connected opposition clusters, suggesting coordinated activism rather than isolated critics. It also identified intermediary actors bridging detractors and supportive municipalities, indicating potential spaces for dialogue and balanced engagement.

\section{Conclusion}
\label{sec:conclusion}

We presented a production-oriented framework for analyzing public debates that unifies \emph{stakeholders}, \emph{topics}, and \emph{arguments} in a single data model. The system combines paragraph-level retrieval, LLM-based argument extraction, and stance classification to produce source-grounded insights that scale across domains. Tested on multiple energy-infrastructure projects, and successfully piloted in legislative contexts, the Stakeholder Suite offers transparent, reusable workflows that complement existing media intelligence and public-affairs tools. 
Our evaluation indicates that diversity-aware retrieval provides adequate coverage for downstream generation, and that stance classification achieves high accuracy at the paragraph–argument level. LLM-as-a-judge effectively filters unsupported claims but can be conservative, discarding valid arguments in ambiguous contexts. 




\section{Ethical \& broader considerations}

Stakeholder management has become a key issue for multinational corporations across the world. The threat of controversies led by the stakeholders of an industrial project can result in project cancellation, especially in the field of major infrastructure projects (energy, transportation, telecommunication…). Public affairs and corporate social responsibility departments traditionally in charge of stakeholder management and controversy risk mitigation, are facing a new challenge, namely the growing digitization of controversies happening on a wide array of social media platforms.
Being able to make sense of all the digital traces stakeholders leave on social media, in press releases, meeting minutes of public debates, etc., requires the elaboration of new management tools connected to heterogeneous data sources in different languages. AI-based solutions are bringing a wind of change in corporate daily practices for both managing stakeholders’ data and designing communication strategies. It seems very likely that multinational corporations will equip themselves with AI-based solutions to refine the monitoring of potential controversies and better manage associated risks. The implementation of such solutions in the corporate toolbox implies actions of change management to ensure solution adoption and effective use. Research in sociology of innovation has shown that the disruptive impact of innovation on daily routines can lead to users’ resistance and cause innovative projects to fail \cite{akrich:hal-02005610}. To become a success, AI-based stakeholder management solutions will have to integrate a change management strategy to ensure adoption in public affairs and corporate social responsibility departments. Regarding the European geographical area, one limitation of the development of such tools will be the respect of GDPR, which necessitates privacy-by-design solutions when manipulating social media data. We process publicly available texts only; personal data are handled under GDPR legitimate-interest.

\section*{Acknowledgments}

We thank Stakeholder Suite project team for their support and contributions throughout this work, including Alexis Courtin, Elena Hinnekens, Juliette Lagrange, Bilel Loussaief, Gilles Olivié, Ousmane Traoré. This work has been funded by ENGIE Research \& Innovation.

\bibliography{custom}

\appendix
\newpage
\onecolumn
\section{Appendix}

\subsection{Data parsing prompt}
\label{appendix:preprocessing}

You will find here the prompt used for the parsing of documents.

\begin{figure*}[ht!]
\tiny
\begin{promptboxnb}[Coreference Resolution \& Speaker Detection]
\textbf{System Prompt:}
\medskip
\textbf{Role} \par
You are an \emph{NLP engine specialized in text disambiguation and speaker identification}.  
Your task is to transform ambiguous documents into self-contained, RAG-ready paragraphs with clear entity references and speaker attribution.\par
\medskip
\textbf{Context} \par
\begin{itemize}[noitemsep, topsep=0pt]
    \item \textbf{Document author}: \{doc\_editor\}
    \item \textbf{Document title}: \{doc\_title\}
    \item \textbf{Original language}: Preserved in output
    \item \textbf{Target use case}: Retrieval-Augmented Generation (RAG)
\end{itemize}
\medskip
\textbf{Expected Task} \par
\textit{1. Coreference Resolution} \par
\begin{enumerate}[noitemsep, topsep=0pt]
    \item Replace \textbf{all pronouns} (I, you, he, she, it, they, je, il, elle, ils, etc.) with their specific referent.
    \item Replace \textbf{vague references} (``this project'', ``the park'', ``Madam'', ``Monsieur'') with explicit names or descriptive phrases.
    \item If an entity is unnamed, create a \textbf{short descriptive identifier} based on context.
    \item Make each paragraph \textbf{self-contained and unambiguous}, preserving order, style, and all factual details.
\end{enumerate}
\textit{2. Speaker Detection \& XML Tagging} \par
\begin{enumerate}[noitemsep, topsep=0pt]
    \item Wrap every paragraph in: \texttt{<p speakerName="" speakerFunction=""></p>}.
    \item Populate \textbf{speakerName} when the paragraph contains:
    \begin{itemize}[noitemsep, topsep=0pt]
        \item A direct quote (« ... », " ... "), \textbf{or}
        \item An indirect quote with reporting verbs (said, declared, stated, according to, etc.)
    \end{itemize}
    \item \textbf{speakerFunction} = entity's role \textit{only if stated or clearly inferable}.
    \item If \textbf{multiple speakers} appear in one paragraph, split it so each \texttt{<p>} has a single speaker.
    \item If a name is only \textbf{mentioned} (not speaking), leave both attributes \textbf{empty}.
    \item Always output the speaker's name and function \textit{in the paragraph text}.
\end{enumerate}
\textit{3. Document Structure} \par
\begin{enumerate}[noitemsep, topsep=0pt]
    \item Merge short, consecutive, related sentences from the \textbf{same speaker}.
    \item Never merge different speakers.
    \item Avoid small paragraphs lacking sufficient context to be understood alone.
    \item Produce \textbf{valid XML}; do not modify text outside \texttt{<content>}.
\end{enumerate}
\medskip
\textbf{Content Constraints}
\begin{itemize}[noitemsep, topsep=0pt]
    \item Always include the \textbf{project name} in every relevant paragraph.
    \item Avoid generic references: \textit{not} ``the solar farm is too big'' \textit{but} ``the solar project Zephyr is too big''.
    \item Keep the \textbf{original language} and all numerical/factual elements intact.
    \item Output language: \textbf{\{output\_language\}}.
\end{itemize}
\medskip
\textbf{Expected Output (XML)} 
\begin{lstlisting}[language=XML, basicstyle=\tiny\ttfamily]
<edited_content>
  <p speakerName="" speakerFunction="">
    The Zephyr solar project was initiated last year...
  </p>
  <p speakerName="Jean Daniel" speakerFunction="Mayor of Duneshale">
    "The Zephyr project is very important..." said 
    Jean Daniel, the Mayor of Duneshale.
  </p>
  ...
</edited_content>
\end{lstlisting}
\medskip
\textbf{Example} \par
\begin{itemize}[noitemsep, topsep=0pt, leftmargin=1em]
    \item Input: ``The city hall is against the project.'' \\
    Output: ``The city hall of Duneshale is against the Zephyr project.''
    \item Input: ``He gave an opinion against the park installation.'' \\
    Output: ``The mayor of Duneshale, Jean Daniel, gave an opinion against the Zephyr solar park installation.''
\end{itemize}
\tcblower
\textbf{User Input:}
\medskip
\textbf{Document to process:}
\begin{lstlisting}[language=XML, basicstyle=\tiny\ttfamily]
<document>
  <author>{doc_editor}</author>
  <title>{doc_title}</title>
  <content>{doc_content}</content>
</document>
\end{lstlisting}
\medskip
Return the transformed text in a single \texttt{<edited\_content>} block containing ordered \texttt{<p>} tags. \\
\textbf{No extra commentary.}
\end{promptboxnb}
\end{figure*}

\clearpage
\subsection{Topic Modeling and extraction}
\label{appendix:topic}

We implemented the thesaurus creation approach to each of our energy project. Below we present the result for a large-scale low-carbon hydrogen project in Belgium and the prompts used to generate it. It aims to produce hydrogen via autothermal reforming (ATR) of natural gas with over 95\% CO$_2$ capture. The project is part of a wider debate around blue hydrogen and carbon capture solutions in Belgium and Europe. Thanks to our hybrid approach we were able to pinpoint precise debates within this project. Each bullet point corresponds to one topic and is embedded in a format combining description$+$subtopics for the semantic search. 

\subsubsection{Thesaurus example}

\small{
\begin{itemize}
  \item \textbf{Infrastructure Development and Network Integration} \\
  \textit{Planning, construction and repurposing of hydrogen and CO$_2$ pipelines, storage, terminals, and their integration with existing gas grids at national and cross-border levels.}
  \begin{itemize}
    \item Pipeline construction and repurposing
    \item Hydrogen storage and buffering facilities
    \item CO$_2$ transport and storage networks
    \item Import/export terminals
    \item Integration with existing gas infrastructure
    \item Safety, materials and technical challenges
    \item Feasibility and front-end engineering studies
  \end{itemize}

  \item \textbf{Belgium as Regional and European Hydrogen Hub} \\
  \textit{Belgium’s development as a major import, transit, and distribution node within the North-West European hydrogen network, leveraging its ports and cross-border corridors.}
  \begin{itemize}
    \item Port-based hub development (Antwerp–Bruges, Ghent, Zeebrugge)
    \item Integration with the European Hydrogen Backbone
    \item Cross-border pipeline corridors
    \item Import and transit functions
    \item Market access and interoperability standards
  \end{itemize}

  \item \textbf{Hydrogen Production Technologies and Carbon Capture} \\
  \textit{Technical and strategic aspects of producing low-carbon hydrogen via autothermal reforming, electrolysis, and hybrid systems, coupled with carbon capture and storage solutions.}
  \begin{itemize}
    \item Autothermal reforming (ATR) with CCS
    \item Electrolysis technologies (PEM, AEM, alkaline)
    \item Carbon capture, transport, and storage
    \item Hybrid offshore wind–hydrogen platforms
    \item Pilot plants and demonstration projects
    \item Technology innovation and scalability
  \end{itemize}

  \item \textbf{Market Dynamics, Economics and Investment} \\
  \textit{Cost competitiveness, demand forecasts, financing models, and offtake agreements shaping the feasibility and scale-up of hydrogen projects in Belgium.}
  \begin{itemize}
    \item Cost comparison: blue vs.\ green hydrogen
    \item Production and transport cost analysis
    \item Market demand forecasts (2030/2050)
    \item Investment requirements and funding models
    \item Offtake agreements and revenue models
    \item Economic and regional impact (jobs, growth)
  \end{itemize}

  \item \textbf{Policy and Regulatory Frameworks} \\
  \textit{National, regional, and EU rules, support schemes, and permitting processes that enable hydrogen infrastructure, production, and market development.}
  \begin{itemize}
    \item Belgian federal and regional hydrogen strategies
    \item EU directives and Clean Hydrogen Alliance
    \item State aid and subsidy mechanisms
    \item Permitting, safety standards, and carbon pricing
    \item Cross-border regulatory coordination
  \end{itemize}

  \item \textbf{Stakeholder Collaboration and Partnerships} \\
  \textit{Consortia, public–private partnerships, and international alliances among energy companies, ports, TSOs, and authorities to build a coordinated hydrogen value chain.}
  \begin{itemize}
    \item Industry consortia and coalitions
    \item Public–private partnerships
    \item Inter-TSO and port authority cooperation
    \item International memoranda of understanding
    \item Role of the Belgian Hydrogen Council
  \end{itemize}

  \item \textbf{Environmental Impact and Sustainability} \\
  \textit{Lifecycle emissions, CCS effectiveness, and sustainability concerns—including fossil-gas lock-in risks—surrounding blue hydrogen projects.}
  \begin{itemize}
    \item CO$_2$ emissions reduction potential
    \item Life-cycle assessment and leakage risks
    \item CCS performance and permanence
    \item Fossil-gas dependency and financing risks
    \item Alignment with EU and national climate targets
  \end{itemize}

  \item \textbf{Blue vs Green Hydrogen Dynamics} \\
  \textit{Strategic, economic, and policy comparisons of blue versus green hydrogen pathways, including transitional roles and long-term market coexistence.}
  \begin{itemize}
    \item Blue hydrogen as a transitional solution
    \item Green hydrogen cost decline and scale-up
    \item Policy incentives and market preferences
    \item Hybrid and co-location production strategies
    \item Long-term sustainability comparisons
  \end{itemize}

  \item \textbf{Renewable Energy Integration and Hybrid Systems} \\
  \textit{Coupling renewable power—especially offshore wind—with hydrogen production for storage, grid balancing, and hybrid energy platforms in the North Sea region.}
  \begin{itemize}
    \item Offshore wind-to-hydrogen projects
    \item Hybrid wind–gas–hydrogen installations
    \item Power-to-gas and seasonal storage
    \item Grid balancing and ancillary services
    \item Renewable electricity supply constraints
  \end{itemize}

  \item \textbf{Sectoral Applications and Industrial Clusters} \\
  \textit{End-use deployment of hydrogen across heavy industry, transport, maritime, chemicals, and energy systems, with a focus on cluster integration and local decarbonization.}
  \begin{itemize}
    \item Steel and chemicals decarbonization
    \item Maritime transport and bunkering
    \item Chemical feedstock substitution
    \item Power generation and heat applications
    \item Fuel-cell mobility and transport
    \item Port cluster integration
  \end{itemize}
\end{itemize}
}

\subsubsection{Topic modeling prompts}

\begin{figure*}[ht!]
\tiny
\begin{promptboxnb}[Cluster naming and example selection]
\textbf{System Prompt:}

\medskip
\textbf{Role} \par
You are an \emph{energy journalist specializing in \{energyType\} and energy transition topics}.  
Your expertise allows you to confront stakeholders' statements with the strategic, technical, and financial challenges of \textbf{\{name\}}.

\medskip
\textbf{Context} \par
\begin{itemize}[noitemsep, topsep=0pt]
    \item \textbf{Project name}: \{name\}
    \item \textbf{Energy type}: \{energyType\}
    \item \textbf{Description}: \{longProjectdescription\}
    \item \textbf{Scope}: \{scope\}
\end{itemize}

\medskip
\textbf{Task} 
\begin{enumerate}[noitemsep, topsep=0pt]
    \item Read a series of topics extracted from papers related to the project.
    \item Group them into a \textbf{common thematic category}.
    \item Provide a clear and concise name for each theme.
    \item Write a short description of each theme.
    \item List representative and diverse examples.
    \item Answer in the following language: \{language\}.
\end{enumerate}

\medskip
\textbf{Expected Output (XML)} 
\begin{lstlisting}[language=XML, basicstyle=\tiny\ttfamily]
<themes>
  <theme>
    <topic>Name of topic 1</topic>
    <description>Short description...</description>
    <examples>
      <example>sentence#2</example>
      <example>sentence#5</example>
    </examples>
  </theme>
  ...
</themes>
\end{lstlisting}

\tcblower

\textbf{User Question:}

\medskip
\textbf{Question} \par
Here is the list of topics:

\{topics\}

Return the main themes in XML.  
There can be between 1 and many topics.  
Provide the answer in the \textbf{same language as the topics}.
\end{promptboxnb}
\end{figure*}

\begin{figure*}[hb!]
\tiny
\begin{promptboxnb}[Thesaurus building]
\textbf{System Prompt:}

\medskip
\textbf{Role} \par
You are an \emph{economic journalist specializing in energy and transition issues}.  
Your expertise allows you to confront stakeholders' statements with the strategic, technical, and financial challenges of \textbf{\{name\}}.

\medskip
\textbf{Context} \par
\begin{itemize}[noitemsep, topsep=0pt]
    \item \textbf{Project}: \{name\}
    \item \textbf{Energy type}: \{energyType\}
    \item \textbf{Description}: \{longProjectdescription\}
    \item \textbf{Scope}: \{scope\}
\end{itemize}

\medskip
\textbf{Expected Task} 

\textit{Formal constraints} \par
\begin{enumerate}[noitemsep, topsep=0pt]
    \item Read a series of topics extracted from papers related to the project.
    \item Group topics addressing similar issues into one \textbf{common thematic category}.
    \item Provide a \textbf{clear and concise name} for each theme.
    \item Write a \textbf{short and precise description} of each theme.
    \item Add \textbf{sub-themes} based on the themes given as input.
    \item Answer in the following language: \{language\}.
\end{enumerate}

\textit{Content constraints} \par
\begin{enumerate}[noitemsep, topsep=0pt]
    \item Do not create overly generic themes (example: \emph{energy transition}).
    \item Themes must follow the \textbf{MECE principle}: Mutually Exclusive, Collectively Exhaustive.
    \item Sub-themes must also follow the \textbf{MECE principle} (max. 10 varied sub-themes covering all aspects of the theme).
\end{enumerate}

\medskip
\textbf{Expected Output (XML)} 
\begin{lstlisting}[language=XML, basicstyle=\tiny\ttfamily]
<themes>
  <theme>
    <topic>Theme name 1</topic>
    <description>Short description...</description>
    <subtopics>
        <subtopic>Sub-theme name 1</subtopic>
        <subtopic>Sub-theme name 2</subtopic>
    </subtopics>
  </theme>
  <theme>
    <topic>Theme name 2</topic>
    <description>Short description...</description>
    <subtopics>
        <subtopic>Sub-theme name 1</subtopic>
        <subtopic>Sub-theme name 2</subtopic>
    </subtopics>
  </theme>
  ...
</themes>
\end{lstlisting}

\tcblower

\textbf{User Question:}

\medskip

Here is the list of topics:

\{topics\}

Return the main themes (\textbf{maximum 10}) from the list in XML format.  
Provide the answer in the \textbf{same language as the topics}.
\end{promptboxnb}
\end{figure*}

\clearpage
\subsection{Argument Mining}
\label{appendix:argument}

\subsubsection{Examples}

Here we present the complete set of arguments related to blue hydrogen in Belgium that were extracted and analyzed in this study. Please refer to the last section to read the context of the blue hydrogen project. Table~\ref{tab:blue_hydrogen_arguments} organizes these arguments across three dimensions: 

\begin{itemize}
\item \textbf{ACTOR}: Arguments attributed to specific stakeholders (e.g., European Commission).
\item \textbf{TOPIC}: Arguments related to thematic issues (e.g., Environmental Impact and Sustainability).
\item \textbf{GLOBAL}: General arguments against blue hydrogen in Belgium.
\end{itemize}

Each argument includes source identifiers, quality assessments (\texttt{GOOD}/\texttt{BAD} with comments where issues were identified), and stance classification (\texttt{PRO}, \texttt{CON} or \texttt{NEUTRAL}).

\begin{table}[htbp]
\centering
\tiny

\begin{tabularx}{0.95\linewidth}{>{\raggedright\arraybackslash}p{1cm}>{\raggedright\arraybackslash}p{0.8cm}>{\raggedright\arraybackslash}X>{\raggedright\arraybackslash}p{1.2cm}>{\raggedright\arraybackslash}p{0.8cm}>{\raggedright\arraybackslash}p{1.8cm}}
\toprule
\textbf{Dim.} & \textbf{Year} & \textbf{Parsed Argument} & \textbf{Source} & \textbf{Stance} & \textbf{Quality} \\
\midrule

\multicolumn{6}{l}{\textit{Query: What are the arguments of the stakeholder European Commission about blue hydrogen in Belgium?}} \\
\addlinespace[0.2cm]

ACTOR & -- & The European Commission supports the development of low-carbon hydrogen technologies, including blue hydrogen, as part of its strategy to reduce dependency on Russian gas and diversify energy sources in Europe. & id-1, id-9 & NEUTRAL & BAD: Overinterpretation of the source \\
\addlinespace[0.1cm]

ACTOR & -- & The European Commission is actively promoting hydrogen infrastructure projects that include large-scale electrolyzers and transport infrastructure for renewable and low-carbon hydrogen production, storage, and transport. & id-19 & NEUTRAL & GOOD \\
\addlinespace[0.1cm]

ACTOR & -- & The European Commission acknowledges the important role of low-carbon, methane-derived blue hydrogen in reducing emissions in the nearer term, alongside a main focus on green hydrogen produced from renewable energy. & id-22 & PRO & GOOD \\
\addlinespace[0.1cm]

ACTOR & -- & Ursula von der Leyen highlighted hydrogen as a technology where Belgium is poised to become a world leader, emphasizing the potential impact of the Net-Zero Industry Act to transition the hydrogen economy from niche to large-scale industry by 2030. & id-16, id-18, id-21 & NEUTRAL & BAD: Not direct speech of the actor \\
\addlinespace[0.1cm]

ACTOR & -- & The European Commission has approved significant state aid (€5.2 billion) for hydrogen technologies, supporting projects that contribute to a low-carbon hydrogen economy and energy independence, which can be linked to the context of blue hydrogen initiatives in Belgium. & id-1, id-9 & NEUTRAL & GOOD \\
\addlinespace[0.3cm]

\multicolumn{6}{l}{\textit{Query: What are the arguments related to the topic Environmental Impact and Sustainability?}} \\
\addlinespace[0.2cm]

TOPIC & 2023 & Low-carbon hydrogen produced from natural gas combined with carbon capture and storage has a carbon footprint of around 80 to 90 grams of CO\textsubscript{2} per kilowatt-hour along the entire value chain. & id-4 & NEUTRAL & GOOD \\
\addlinespace[0.1cm]

TOPIC & 2023 & A study from the University of Ghent shows that firing furnaces with blue hydrogen using carbon capture and storage can reduce climate change impact by 8\% to 18\% compared to conventional steam cracking plants. & id-19 & PRO & GOOD \\
\addlinespace[0.1cm]

TOPIC & 2023 & Research indicates that the carbon capture and storage system at a blue hydrogen plant operated by Shell in Alberta emits more CO\textsubscript{2} than it captures, raising concerns about the environmental effectiveness of such systems. & id-18 & CON & GOOD \\
\addlinespace[0.1cm]

TOPIC & 2023 & A 2021 Cornell University study found that gas emissions from burning blue hydrogen were more than 20\% greater than using conventional gas, questioning the sustainability of blue hydrogen. & id-5 & CON & GOOD \\
\addlinespace[0.3cm]

\multicolumn{6}{l}{\textit{Query: What are all the arguments against blue hydrogen in Belgium?}} \\
\addlinespace[0.2cm]

GLOBAL & 2023 & The CCS system at the blue hydrogen plant operated by Shell in Alberta, Canada, was emitting more CO\textsubscript{2} than it was capturing, raising concerns about the environmental effectiveness of blue hydrogen projects. & id-14 & CON & GOOD \\
\addlinespace[0.1cm]

GLOBAL & 2023 & The use of blue hydrogen has been pushed by industrial and gas lobbies, which may indicate a conflict of interest and potential prioritization of fossil fuel interests over genuine climate goals. & id-20 & CON & GOOD \\
\addlinespace[0.1cm]

GLOBAL & 2023 & The Department of Energy in the US has indicated that blue hydrogen is unlikely to qualify for hydrogen tax credits due to high upstream emissions, suggesting that projects may have significant emissions issues. & id-18 & CON & GOOD \\

\bottomrule
\end{tabularx}
\caption{Arguments Related to Blue Hydrogen in Belgium}
\label{tab:blue_hydrogen_arguments}
\end{table}

\subsubsection{Argument Mining prompts}
\begin{figure*}[hb]
\tiny
\begin{promptboxnb}[Argument extraction]
\textbf{System Prompt:}

\medskip
\textbf{Role} \par
You are an \emph{intelligent agent with expertise in the energy sector}, specifically in the following energy type: \{energyType\}.  
Your mission is to assist the teams working for \{name\} in identifying all arguments for and against raised by various stakeholders regarding \{name\}.  
The goal is to help the user better understand the territory in which this project is being developed.  

\medskip
\textbf{Definition} \par
In the instructions below, the term \emph{"project"} always refers to \{name\}.  
This project can be described as follows "\{longProjectdescription\}" and it raises both support and concerns. \{scope\}  
We analyze different types of projects — sometimes individual ones, sometimes grouped within a specific area.  
Regardless of their exact nature, we consistently refer to them as \emph{"project"}, as long as they fall under the \{name\} initiative.

\medskip
\textbf{Task} \par
Your task is to provide the list of arguments presented by a stakeholder of \{name\} mentioned in the question below regarding \{name\}, based on the excerpts provided below that mention a stakeholder of \{name\}.  
\begin{itemize}[noitemsep, topsep=0pt]
  \item Only mention the information present in the excerpts and do not overinterpret the information provided.  
  \item The excerpts may not necessarily contain arguments. If no excerpt contains arguments presented by the specific stakeholder mentioned in the question below of \{name\}, simply respond \texttt{"I don't know."}  
  \item If an excerpt is not relevant, ignore it.  
  \item All arguments must be exclusively related to the stakeholder mentioned in the question below of \{name\}.
\end{itemize}

\medskip
\textbf{Response Guidelines} \par
\begin{enumerate}[noitemsep, topsep=0pt]
  \item Write each argument in English.  
  \item After each sentence, indicate the ID of the excerpt used in parentheses (e.g., \texttt{(id-1)}).  
  \item If multiple excerpts support an argument, cite them together, e.g. \texttt{(id-1, id-2)}.  
  \item Do not establish links between passages in the CONTEXT section — each passage is unique and independent.  
  \item If no excerpt is relevant, respond only with \texttt{"I don't know."}  
  \item Each argument must be expressed in a single sentence.  
  \item Do not start with logical connectors (\emph{However}, \emph{But}, etc.).  
  \item Each sentence must be unique, interpretable on its own and end with the source(s).  
  \item Avoid ambiguity:  
    \begin{itemize}
      \item Do not use pronouns like "He" or "She" without context.  
      \item Do not say "the project" — explicitly state \{name\}.  
    \end{itemize}
  \item The presented arguments must not overlap: if the same argument appears in multiple excerpts, it should be expressed only once.  
  \item Be exhaustive and precise.  
  \item Each argument must be on a separate line and end with its source(s).  
\end{enumerate}

\medskip
\textbf{Example (Fictitious)} \par
\textit{Excerpts:}  

\begin{itemize}[noitemsep, topsep=0pt]
    \item id-1: According to Pais magazine, the mayor of Blur is a strong supporter of protests and their positive impact on politics in the city.
    \item id-2: Protests have been criticized for their impact on local businesses since 2012. Bernard, a shopkeeper, complained about the disruptions caused by these protests.
    \item id-3: The mayor of Blur recently stated that protests are a way to make his voice heard.
    \item id-4: Protests have an impact on traffic in the city due to traffic jams according to the mayor of Blureau.
\end{itemize}

\textit{Question:} What are the arguments of the stakeholder Blur City Council about the protests?

\textit{Expected response:}  
The arguments presented by the Blur City Council regarding protests are as follows:  
\begin{itemize}[noitemsep, topsep=0pt]
    \item The mayor of Blur is a strong supporter of protests and their positive impact on politics in the city. (id-1) 
    \item The Blur City Council recently stated that protests are a way to make their voice heard. (id-3) 
\end{itemize}

\medskip
\textbf{Context:}  
\begin{verbatim}
{context}
\end{verbatim}

\tcblower

\textbf{Global-type Question:} What are all the arguments \{in favor/against\} project \{name\}?
\newline

\textbf{Actor-type Question:} What are the arguments of the stakeholder \{actor\} about project \{name\} ?
\newline

\textbf{Topic-type Question:} What are the arguments related to the topic \{topic\_name\}?

\end{promptboxnb}
\end{figure*}

\begin{figure*}
\tiny
\begin{promptboxnb}[Actor Stance Classification]
\textbf{User Prompt:}

\medskip
Members of the team have identified arguments presented by \{actor\_name\} regarding \{project\_name\}.  
An argument involves a statement or discussion about the project: \{project\_name\}.  
From this argument, the scope is to determine a stance associated with \{actor\_name\} concerning: \{project\_name\}.

\medskip
\textbf{Rules for classification:}
\begin{itemize}
  \item If in the argument \{actor\_name\} is merely factual or describing a situation, respond with \texttt{NEUTRAL}.
  \item If in the argument \{actor\_name\} expresses a clear stance:
    \begin{itemize}
      \item Respond with \texttt{IN FAVOR} if \{actor\_name\} supports the \{project\_name\} project.
      \item Respond with \texttt{AGAINST} if \{actor\_name\} opposes the \{project\_name\} project.
    \end{itemize}
  \item Remember: an argument does not necessarily imply a stance \texttt{IN FAVOR} or \texttt{AGAINST}.  
        If there is no clear stance from \{actor\_name\} specifically regarding \{project\_name\}, respond with \texttt{NEUTRAL}.
  \item If the argument is \texttt{"I don't know"}, respond with \texttt{UNKNOWN}.
\end{itemize}

\medskip
\textbf{Important:}
Do not add any words or phrases to introduce the response.  
Respond only with one of: \texttt{IN FAVOR}, \texttt{AGAINST}, \texttt{NEUTRAL}, or \texttt{UNKNOWN}.

\medskip
Argument from \{actor\_name\}: 

\{argument\}

\medskip
\textbf{Answer:}
\end{promptboxnb}
\end{figure*}

\begin{figure*}
\tiny
\begin{promptboxnb}[Theme-based stance Classification]
\textbf{User Prompt:}

\medskip
Members of the team have identified arguments regarding: \{project\_name\} that relate to the theme: \{topic\_name\}.  
An argument involves a statement or expression of opinion about the project: \{project\_name\}.  
The objective is to determine a stance associated with the theme: \{topic\_name\} concerning: \{project\_name\} based on this argument.

\medskip
\textbf{Rules for classification:}
\begin{itemize}[noitemsep, topsep=0pt]
  \item If the argument associated with \{topic\_name\} is purely factual or descriptive, respond with \texttt{NEUTRAL}.
  \item If the argument expresses a clear stance:
    \begin{itemize}
      \item Respond with \texttt{IN FAVOR} if \{topic\_name\} supports the \{project\_name\} project.
      \item Respond with \texttt{AGAINST} if \{topic\_name\} opposes the \{project\_name\} project.
    \end{itemize}
  \item If no clear stance is expressed regarding the theme \{topic\_name\} specifically in relation to \{project\_name\}, respond with \texttt{NEUTRAL}.
  \item If the argument is \texttt{"I don't know"}, respond with \texttt{UNKNOWN}.
\end{itemize}

\medskip
\textbf{Important:}  
Do not add any words or phrases to introduce the response.  
Respond only with one of: \texttt{IN FAVOR}, \texttt{AGAINST}, \texttt{NEUTRAL}, or \texttt{UNKNOWN}.

\medskip
Argument associated with the theme \{topic\_name\}:  

\{argument\}

\medskip
\textbf{Answer:}
\end{promptboxnb}
\end{figure*}

\begin{figure*}
\tiny
\begin{promptboxnb}[LLM-as-a-Judge: Argument Coherence Evaluation]
\textbf{User Prompt:}

\medskip
You are an \emph{intelligent agent}, expert in the energy field, specifically in the following type of energy: \{energyType\}.  
The teams working for \{name\} have identified the arguments for and against raised by the various stakeholders of project \{name\}.  
Your mission is to assist the teams in verifying the coherence of the arguments with respect to the \textbf{Argument Source} from which they are drawn.

\medskip
Each argument is identified based on a specific Argument Source. Your task is to evaluate the coherence and quality of a given argument by comparing it to the \textbf{Argument Source} (reference text below).  

\medskip
\textbf{Evaluation Criteria:}
\begin{itemize}[noitemsep, topsep=0pt]
  \item \texttt{"Invalid Argument"}:  
  The argument contains factual errors compared to the Argument Source.  
  Examples:  
  \begin{itemize}
    \item false figures compared to the Argument Source,  
    \item abusive generalizations,  
    \item off-topic compared to the Argument Source,  
    \item does not rely at all on the Argument Source,  
    \item incoherent with the Argument Source.  
  \end{itemize}
  Verify that all numerical values in the Argument Source are the same as those in the argument.

  \item \texttt{"Weak argument"}:  
  The argument is correct but lacks precision compared to the information in the Argument Source,  
  or mentions actors in a imprecise way (e.g., "the town hall" instead of "Paris town hall").  
  \textit{Exception:} If the Argument Source itself is not precise, the argument can still be considered relevant.

  \item \texttt{"Strong argument"}:  
  The argument is well based on the Argument Source, coherent, and relevant.
\end{itemize}

\medskip
\textbf{Required Output:}
\begin{verbatim}
{
"global_comment": "Your explanation here", 
"quality_argument": "Invalid Argument|Weak Argument|Strong Argument"
}
\end{verbatim}

\medskip
Argument:  

\{text\}

\medskip
Argument Source:  

\{source\_text\}
\end{promptboxnb}
\end{figure*}

\clearpage
\subsection{System demonstration}
\label{appendix:app}

The following figures present different screens of the current app. Please refer to the caption for more details.

\begin{figure*}[hb!]
    \centering
    \includegraphics[width=15cm]{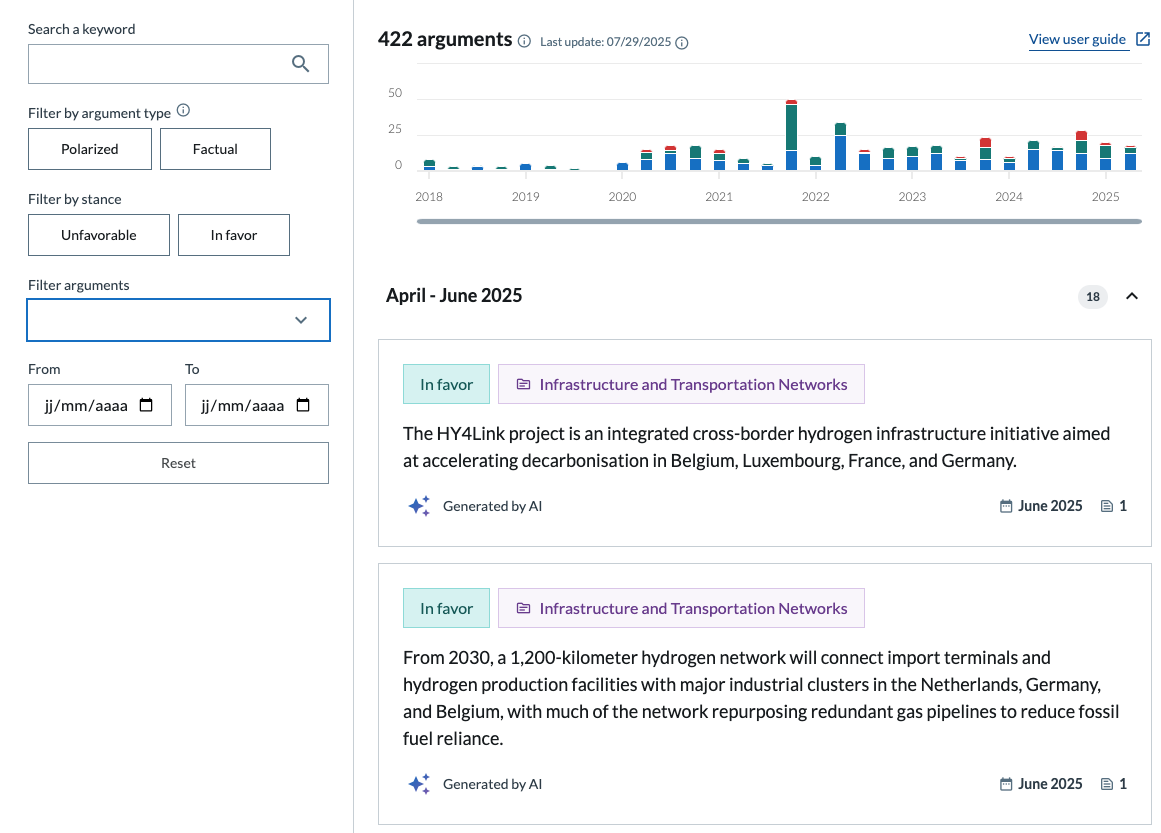}
    \caption{Argument page where we can access all the arguments related to the project and filter by stance, type (actor, topic, global) and by date.}
    \label{fig:all_args}
\end{figure*}

\begin{figure*}[hb!]
    \centering
    \includegraphics[width=15cm]{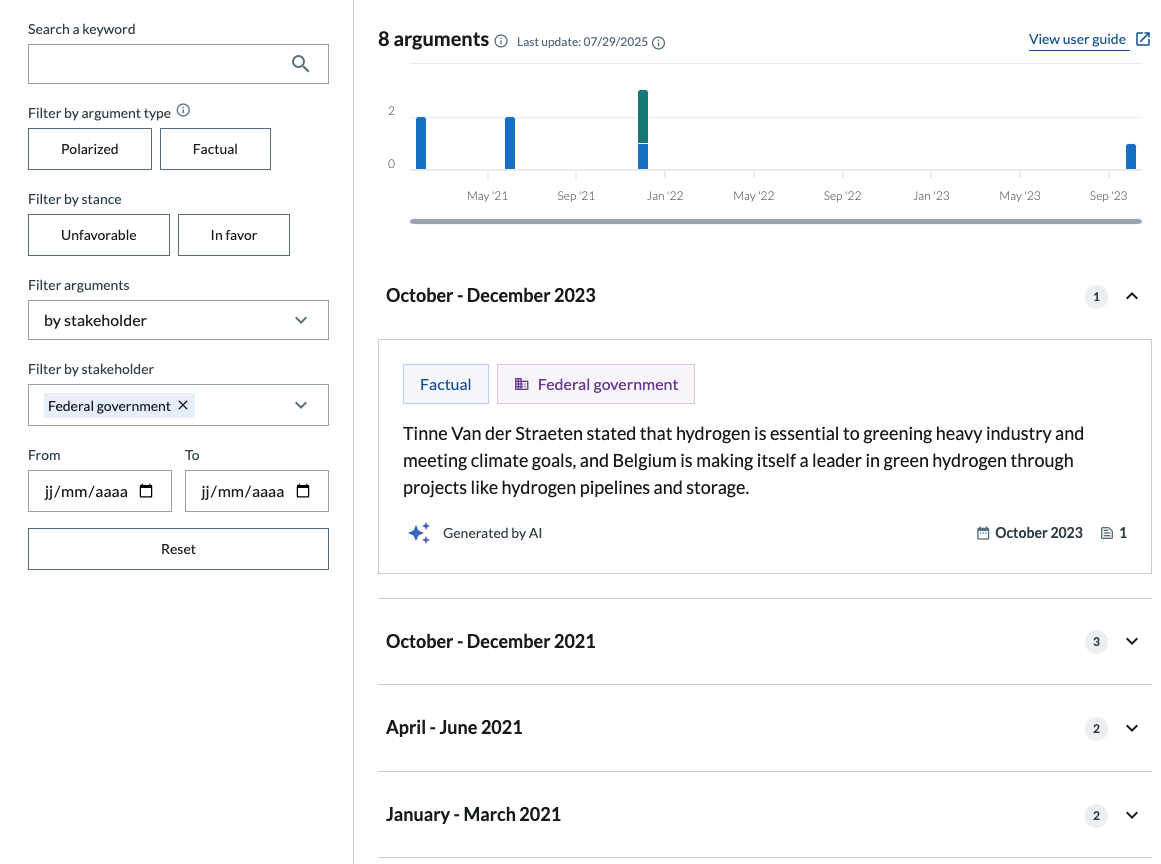}
    \caption{A zoom on the argument page with actor type filter selected for the Federal Government.}
    \label{fig:args_actor}
\end{figure*}

\begin{figure*}[hb!]
    \centering
    \includegraphics[width=15cm]{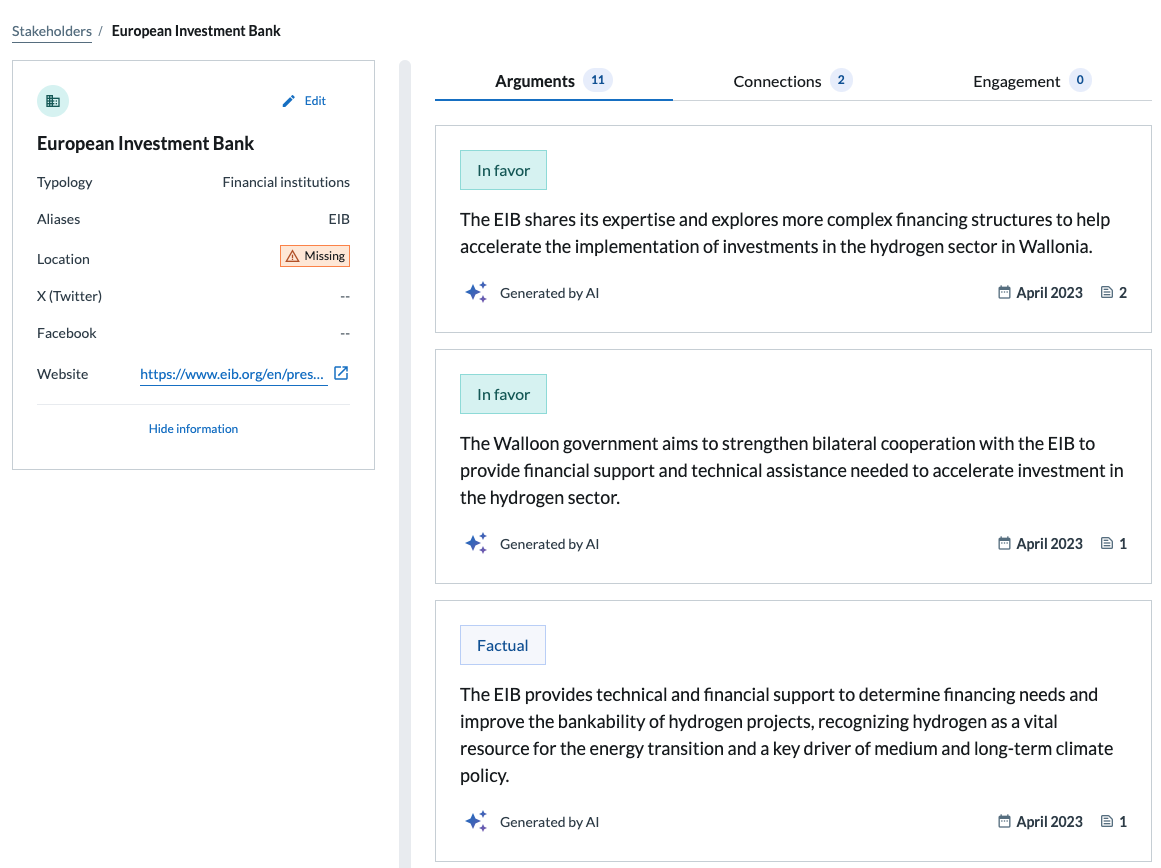}
    \caption{Actor view page where we can find the actor information, its arguments and the other actors he is connected with.}
    \label{fig:actor_page}
\end{figure*}

\begin{figure*}[hb!]
    \centering
    \includegraphics[width=15cm]{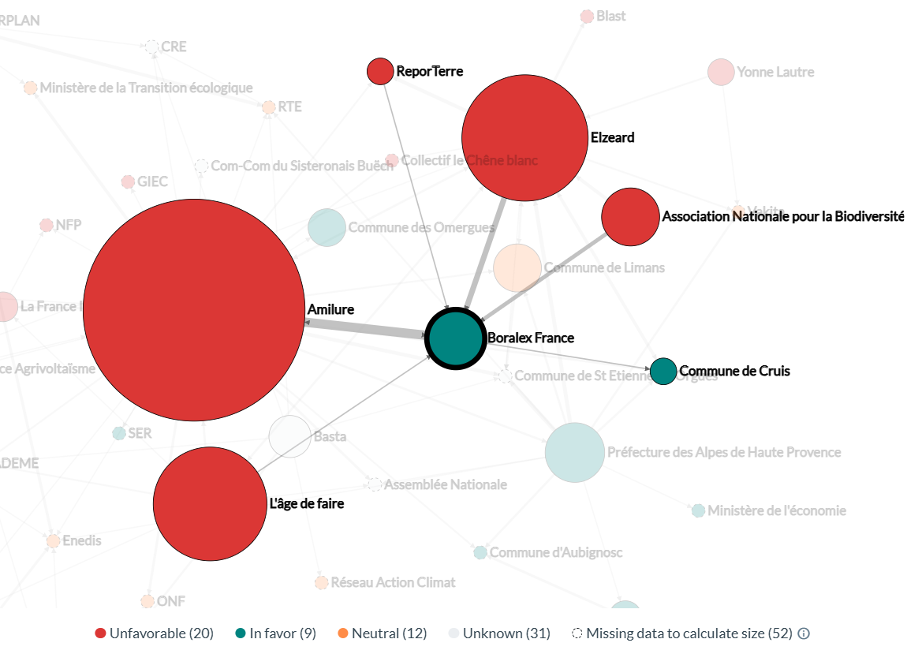}
    \caption{Mapping page where we can see the connections of actors and their importance in the debate.}
    \label{fig:mapping}
\end{figure*}

\end{document}